\documentclass{article}

\usepackage{microtype}
\usepackage{graphicx}
\usepackage{subfigure}
\usepackage{booktabs} 

\usepackage{hyperref}

\usepackage[accepted]{icml2025}

\usepackage{amsmath}
\usepackage{amssymb}
\usepackage{mathtools}
\usepackage{amsthm}

\usepackage[capitalize,noabbrev]{cleveref}

\theoremstyle{plain}

\theoremstyle{definition}

\theoremstyle{remark}

\usepackage[textsize=tiny]{todonotes}

\usepackage{bm}
\usepackage{multirow}
\usepackage{xcolor} 
\usepackage{xspace}

\newcommand*\diffop{\mathop{}\!\mathrm{d}}

\usepackage{array}
\usepackage{subcaption}
\usepackage{multirow,multicol}
\usepackage[flushleft]{threeparttable}
\usepackage{comment}
\usepackage{algorithmic}
\usepackage{booktabs}
\usepackage{bbm, dsfont}
\usepackage[font=small,labelfont=bf]{caption}

\usepackage{amssymb}
\usepackage{pifont}

\newcolumntype{x}[1]{>{\centering\arraybackslash}p{#1pt}}
\newcommand{\app}{\raise.17ex\hbox{$\scriptstyle\sim$}}

\newlength\savewidth

\makeatletter\renewcommand\paragraph{\@startsection{paragraph}{4}{\z@}
  {.5em \@plus1ex \@minus.2ex}{-.5em}{\normalfont\normalsize\bfseries}}\makeatother

\def\tablecite#1#{%
  \def\pretablecite{#1}%
  \tableciteaux}
\def\tableciteaux#1{%
  \textsuperscript{\expandafter\originalcite\pretablecite{#1}}%
}

\usepackage{graphicx}
\usepackage{enumitem}
\usepackage{wrapfig}
\usepackage{lipsum}
\usepackage{soul}

\usepackage{color, colortbl}

\usepackage{tabu}
\usepackage{xcolor}
\usepackage{nicematrix}

\definecolor{ForestGreen}{rgb}{0.13, 0.55, 0.13}
\definecolor{Green}{rgb}{0.0, 0.5, 0.0}
\definecolor{Blue}{rgb}{0.25, 0.42, 0.88}
\definecolor{green(munsell)}{rgb}{0.0, 0.66, 0.47}
\definecolor{green(ryb)}{rgb}{0.4, 0.69, 0.2}
\definecolor{green(pigment)}{rgb}{0.0, 0.65, 0.31}
\definecolor{citecolor}{HTML}{0071bc}
\definecolor{GrayXMark}{gray}{0.7}
\definecolor{DifferenceColor}{HTML}{af3235}
\definecolor{HighlightColor}{gray}{0.9}
\definecolor{OracleTextColor}{gray}{0.55}
\definecolor{Cerulean}{HTML}{00a2e3}

\newcolumntype{H}{>{\setbox0=\hbox\bgroup}c<{\egroup}@{}}
\newcolumntype{a}{>{\columncolor{HighlightColor}}c}
\newcolumntype{L}[1]{>{\centering\arraybackslash}m{#1}}

\usepackage{tabularx}
\usepackage[export]{adjustbox}
\usepackage{makecell}
\usepackage{ragged2e}

\newcommand{\Ours}{DiTo\xspace}

\newcommand{\gae}{GLPTo\xspace}

\newcommand{\daeB}{\Ours-B\xspace}
\newcommand{\daeL}{\Ours-L\xspace}
\newcommand{\daeXL}{\Ours-XL\xspace}

\newcommand{\imnet}{ImageNet\xspace}

\makeatletter
\DeclareRobustCommand\onedot{\futurelet\@let@token\@onedot}
\def\@onedot{\ifx\@let@token.\else.\null\fi\xspace}
\def\eg{\emph{e.g}\onedot} 
\def\ie{\emph{i.e}\onedot} 
 
 \def\vs{\emph{vs}\onedot}

\def\etal{\emph{et al}\onedot}

\icmltitlerunning{}

\begin{document}

\twocolumn[
\icmltitle{Diffusion Autoencoders are Scalable Image Tokenizers}



\icmlsetsymbol{equal}{*}

\begin{icmlauthorlist}
\icmlauthor{Yinbo Chen}{ucsd}
\icmlauthor{Rohit Girdhar}{meta}
\icmlauthor{Xiaolong Wang}{ucsd}
\icmlauthor{Sai Saketh Rambhatla}{meta}
\icmlauthor{Ishan Misra}{meta}
\end{icmlauthorlist}

\icmlaffiliation{ucsd}{UC San Diego}
\icmlaffiliation{meta}{GenAI, Meta}
\icmlcorrespondingauthor{Yinbo Chen}{yic026@ucsd.edu}

\icmlkeywords{Machine Learning, ICML}

\vskip 0.3in
]



\printAffiliationsAndNotice{}  

\begin{abstract}
Tokenizing images into compact visual representations is a key step in learning efficient and high-quality image generative models.
We present a simple diffusion tokenizer (\Ours) that learns compact visual representations for image generation models.
Our key insight is that a single learning objective, diffusion L2 loss, can be used for training scalable image tokenizers.
Since diffusion is already widely used for image generation, our insight greatly simplifies training such tokenizers.
In contrast, current state-of-the-art tokenizers rely on an empirically found combination of heuristics and losses, thus requiring a complex training recipe that relies on non-trivially balancing different losses and pretrained supervised models.
We show design decisions, along with theoretical grounding, that enable us to scale \Ours for learning competitive image representations.
Our results show that \Ours is a simpler, scalable, and self-supervised alternative to the current state-of-the-art image tokenizer which is supervised.
\Ours achieves competitive or better quality than state-of-the-art in image reconstruction and downstream image generation tasks.
Project page and code: \url{https://yinboc.github.io/dito/}.
\vspace{-1em}
\end{abstract}
\section{Introduction}
\label{sec:intro}

\begin{figure}
    \centering
    \includegraphics[width=\linewidth]{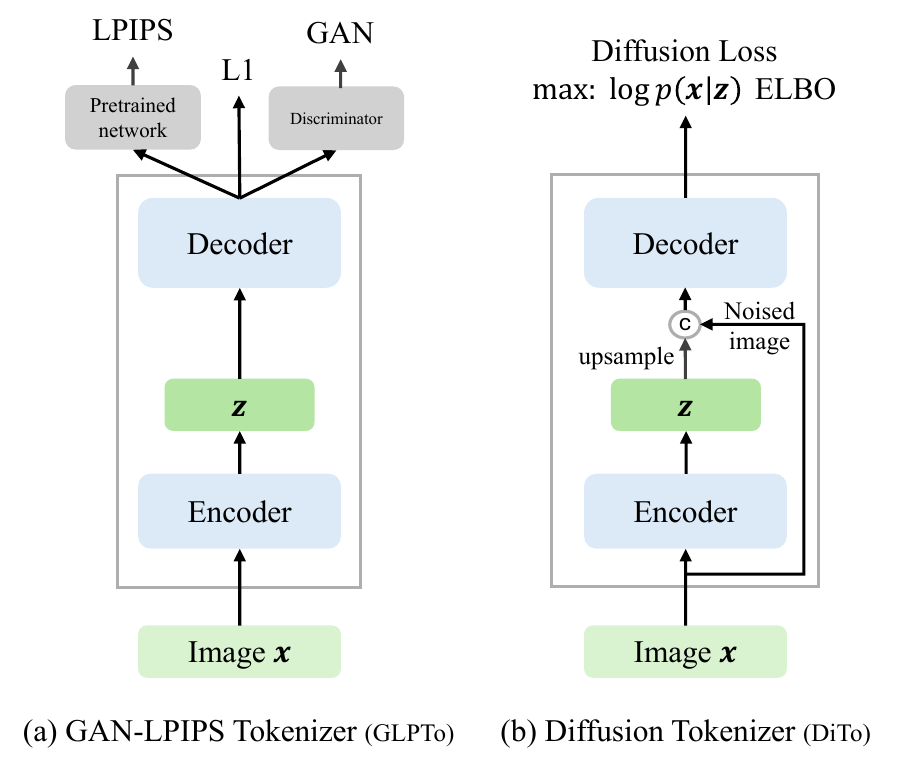}
    \vspace{-2.2em}
    \caption{
    \textbf{Diffusion tokenizer (DiTo)} is a diffusion autoencoder with an ELBO objective (\eg, Flow Matching).
    The input image $\bm{x}$ is passed into the encoder $E$ to obtain the latent representation, \ie, `tokens' $\bm{z}$, a decoder $D$ then learns the distribution $p(\bm{x}|\bm{z})$ with the diffusion objective. $E$ and $D$ are jointly trained from scratch. In contrast, prior work (a) relies on a combination of losses, heuristics, and pretrained models to learn.
    }
    \vspace{-1.3em}
    \label{fig:method}
\end{figure}

Image representations play an important role in the visual generative modeling of images and videos~\cite{rombach2022high,podell2024sdxl,yu2022scaling,dai2023emu,girdhar2023emu,blattmann2023align}.
Since visual data is high dimensional, a dominant paradigm for generative visual models is to first compress the input pixel space into a compact latent representation, then perform generative modeling in the latent space~\cite{rombach2022high,yu2022scaling}, and finally decompress the latent space back to pixel space.
These compact latents have both theoretical and practical benefits.
Compact latents make the generative task easier as the lower dimensional representations remove nuisance factors of variation often present in the raw input signal.
The latents also allow for smaller generative models yielding both training and inference speed-ups.

We focus on the `tokenizers' used to learn the latent representations (tokens) for image generation.
We study the tokenizers commonly used in state-of-the-art image generation methods~\cite{rombach2022high,peebles2023scalable,karras2024analyzing,podell2024sdxl}, which compress the images into continuous latent variables that are further used for learning a latent diffusion generative model.
The image reconstruction quality of the tokenizers directly affects the quality of the generative model and thus, studying and improving the tokenizers is of increasing importance.

\begin{figure*}[!t]
\centering
  \includegraphics[width=\linewidth]{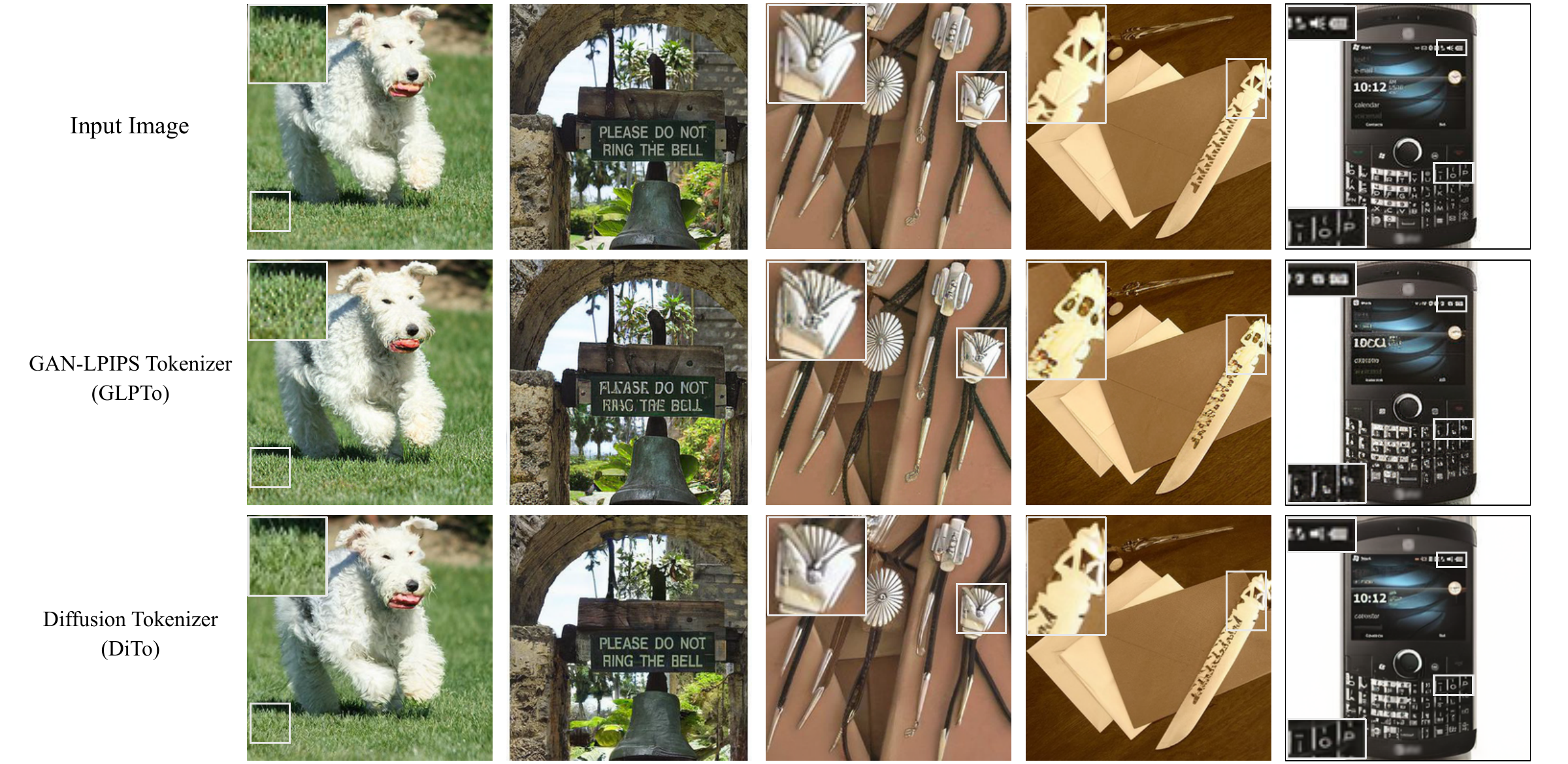}
\captionof{figure}{\textbf{Comparison of GAN-LPIPS tokenizer (\gae) and diffusion tokenizer (\Ours).} \gae uses a weighted combination of L1, LPIPS, and GAN loss, while \Ours only uses a diffusion L2 loss. Despite the simplicity, we observe that when being scaled up, \Ours is competitive to or better than \gae for reconstruction, as shown in the examples (at 256 pixel resolution).}
\label{fig:teaser}
\end{figure*}

The most widely used tokenizer, GAN-LPIPS tokenizer (\gae)~\cite{rombach2022high,peebles2023scalable,karras2024analyzing,podell2024sdxl}, can be viewed as a supervised autoencoder that uses a combination of losses - L1, LPIPS~\cite{zhang2018perceptual} (supervised), and GAN~\cite{goodfellow2020generative} to reconstruct the image (see~\cref{fig:method}).
While effective, \gae is not ideal yet: (i) the combination of several losses requires tuning weights for each of the individual losses; (ii) L1 and LPIPS losses do not correctly model a probabilistic reconstruction, while it is non-trivial to scale up GANs; and (iii) the LPIPS loss is a heuristic that requires a supervised deep network feature space for image reconstruction.
In practice, the \gae reconstructions are prone to have artifacts for structured visual input \eg, text and symbols, and high-frequency image regions as shown in~\cref{fig:teaser}.
These artifacts translate into the image generation model learned on this latent space~\cite{rombach2022high,chen2024image}.
Inspired by these observations, we ask the question: does the image tokenizer training have to be so complex and rely on supervised models?

Diffusion models are a theoretically sound~\cite{kingma2024understanding,dhariwal2021diffusion} and practically scalable~\cite{podell2024sdxl,polyak2024movie} technique for probabilistic modeling of images.
However, the theory and practice of using them for learning representations useful for image generation remains underexplored.
In this work, we show that a single diffusion loss can be used to build scalable image tokenizers.
Our `\textbf{Di}ffusion \textbf{To}kenizer' (\Ours), illustrated in~\cref{fig:method}, is trained with a single diffusion L2 loss.
At inference, given the latent $\bm{z}$, the decoder reconstructs the image from the latent with a diffusion sampler.

We show design choices that allow us to train and scale \Ours yielding competitive or better representations than the \gae.
We connect our training to the recent Evidence Lower Bound (ELBO) theory~\cite{kingma2024understanding} of diffusion models, and use an ELBO objective (Flow Matching~\cite{lipman2023flow}) for the diffusion decoder which makes our learned representations maximize the ELBO of the likelihood of the input image, for which we observe the practical benefits.
Furthermore, we propose noise synchronization, which aims to synchronize the noising process in the latent space to the pixels space, and allows \Ours's latent representation to be more useful for downstream image generation models.

Beyond its simplicity, \Ours achieves competitive or better quality than \gae for image reconstruction, especially for small text, symbols, and structured visual parts.
We also find that image generation models trained on \Ours latent representations are competitive to or outperform those trained on \gae representations.
\Ours can easily be scaled up by increasing the size of the model without requiring any further tuning of loss hyperparameters.
We find both the visual quality and reconstruction faithfulness to the input image get significantly improved when scaling up the model.
Our ablations further suggest that the effectiveness of \Ours lies in jointly learning a latent representation and a decoder for probabilistic reconstruction.

\vspace{-0.5em}
\section{Related Work}

\paragraph{Diffusion models.} Diffusion models are initially proposed and derived as maximizing the evidence lower-bound (ELBO) of data-likelihood in the early work~\cite{sohl2015deep}. Later works~\cite{nichol2021improved,karras2022elucidating} improve various aspects of the initial diffusion model, including architecture, noise schedule, prediction type, and timestep weighting, and connect the theory to score-based generative models~\cite{song2019generative,song2021scorebased}, making many of them no longer follow the derivation in the initial work. When being scaled-up, diffusion models beat GANs for image synthesis~\cite{dhariwal2021diffusion}, and achieve success for various probabilistic modeling tasks, in particular for text-to-image~\cite{nichol2021glide,ramesh2022hierarchical,rombach2022high,betker2023improving,podell2024sdxl} and text-to-video~\cite{girdhar2023emu}. The sampling of diffusion models requires iterative denoising, recent efforts are made towards a faster sampler~\cite{song2021denoising,lu2022dpm} or distilling the diffusion model to a one-step generator~\cite{song2024improved,yin2024onestep,xie2024distillation,salimans2024multistep,yin2024improved,lu2024simplifying}. The recently proposed flow matching~\cite{lipman2023flow} can be also viewed as a diffusion process with a specific simple noise schedule and $\bm{v}$-prediction~\cite{salimans2022progressive} as the training objective.

\paragraph{Image tokenizers.} Image tokenizers are autoencoders that convert images to latent representations that can be reconstructed back. Generative models are then usually trained on the latent representations, including autoregressive models for discrete latents~\cite{esser2021taming}, and diffusion models (or autoregressive diffusion~\cite{li2024autoregressive}) on continuous latents~\cite{rombach2022high}. While diffusion tokenizer is applicable to both types of latents, we focus on continuous latents in this work. A continuous latent space is commonly used by recent state-of-the-art visual generative models~\cite{peebles2023scalable,karras2024analyzing,podell2024sdxl}, which is obtained by a GAN-LPIPS tokenizer~\cite{rombach2022high} (\gae). It uses a combination of L1, LPIPS~\cite{zhang2018perceptual}, and GAN~\cite{goodfellow2020generative} loss for image reconstruction, which is an empirical recipe for reconstruction that is also commonly used in super-resolution~\cite{ledig2017photo,wang2018esrgan,wang2021real}. After obtaining the latent space, a latent diffusion model can be trained with UNet~\cite{rombach2022high} or Transformer~\cite{peebles2023scalable}.

\paragraph{Diffusion autoencoders.} The use of a diffusion objective for training image tokenizers remains largely underexplored.
Early works~\cite{preechakul2022diffusion,pandey2022diffusevae} jointly train an encoder and a diffusion decoder to represent an image as a single latent vector and a noise map for reconstruction. Promising results are shown on simple datasets, while the diffusion autoencoders are mainly used for face attribute editing, 
and they were not connected to the ELBO objectives in recent work~\cite{kingma2024understanding}.
DALL-E 3~\cite{betker2023improving} trains a diffusion decoder to decode from the frozen latent space of the \gae, and distill the diffusion decoder to one-step with consistency model~\cite{song2024improved} for efficiency. Würstchen~\cite{pernias2024wrstchen} trains a diffusion autoencoder to further compress the frozen latent space of a \gae. Concurrent to our work, SWYCC~\cite{birodkar2024samplecompress} uses a diffusion model to refine a coarse prediction supervised by LPIPS loss in a joint training. $\bm{\epsilon}$-VAE~\cite{zhao2024epsilonvaedenoisingvisualdecoding} trains the autoencoder with LPIPS, GAN, and diffusion loss. Both works show that diffusion loss can be helpful in autoencoder training.

\paragraph{Self-supervised representation learning.}
Our work is also related to the research in self-supervised representation learning~\cite{he2020momentum,chen2020simple,misra2020self,he2021masked,caron2021emerging,oord2018representation,donahue2017adversarial,grill2020bootstrap,bao2021beit}. 
In particular, our work leverages the long line of research into methods that leverage an autoencoder style reconstruction loss~\cite{masci2011stacked,ranzato2007unsupervised,vincent2008extracting,he2021masked,salakhutdinov2009deep,bao2021beit}.
While many of these methods are focused on representation learning for downstream recognition tasks, we focus on downstream generation tasks.
We believe studying unified representations for both generation and recognition is a strong research direction for the future.

\vspace{-0.5em}
\section{Preliminaries}
\label{sec:preliminaries}

\paragraph{Score-based models.} Most of the recent state-of-the-art diffusion models are based on the theory of score-based generative models~\cite{song2021scorebased}. A diffusion process~\cite{sohl2015deep,ho2020denoising} gradually adds noise to data and finally makes it indistinguishable from pure Gaussian noise. Formally, given a $D$-dimensional random variable $\bm{x}_0\in \mathbb{R}^D$ that represents the data, the noise schedule is defined by $\alpha_t, \sigma_t$, such that
\begin{align}
    q(\bm{x}_t|\bm{x}_0) = \mathcal{N}(\alpha_t \bm{x}_0, \sigma_t^2 \bm{I}), \quad t\in [0, 1].
\end{align}
A typical design is to let $\alpha_t$ decrease from $\alpha_0=1$ to $\alpha_1=0$, and let $\sigma_t$ increase from $\sigma_0=0$ to $\sigma_1=1$, so that $\bm{x}_{1}\sim \mathcal{N}(\bm{0}, \bm{I})$ is a standard normal distribution.

Diffusion models learn to estimate the score function $\nabla_{\bm{x}} \log q(\bm{x}_t)$~\cite{ho2020denoising,song2020score} for all noise levels $t$.
To estimate the score function, a neural network $\bm{\epsilon}_\theta(\bm{x}_t, t)$ is trained typically with the denoising score matching objective~\cite{ho2020denoising}
\begin{align}
    \mathcal{L}(\bm{x}_0) = \mathbb{E}_{t,\bm{\epsilon}} \big[ ||\bm{\epsilon}_\theta(\bm{x}_t, t) - \bm{\epsilon}||_2^2 \big],
\end{align}
where $\bm{\epsilon}\sim \mathcal{N}(\bm{0}, \bm{I})$, $\bm{x}_t = \alpha_t \bm{x}_0 + \sigma_t \bm{\epsilon}$. After training, $\nabla_{\bm{x}} \log q(\bm{x}_t)\approx -\bm{\epsilon}_\theta(\bm{x}_t, t)/\sigma_t$. A sample of $\bm{x}_0$ can be generated by first sampling $\bm{x}_1$ and then iteratively reversing the diffusion process with the estimated score function using an SDE or ODE solver.

\paragraph{Connection to ELBO.}
The original diffusion loss~\cite{sohl2015deep} is derived by maximizing the evidence lower bound (ELBO) of the log-likelihood of data.
In practice, later works~\cite{ho2020denoising,nichol2021improved,karras2022elucidating} modified the implementation including noise schedule, prediction type, and timestep weighting for improving the visual quality.

These modifications can be viewed as reweighting the loss for denoising tasks at different log signal-to-noise ratios (SNR) $\lambda_t=\log(\alpha_t^2/\sigma_t^2)$:
\begin{align}
    \mathcal{L}(\bm{x}_0) = \frac{1}{2} \int_{\lambda} w(\lambda) \mathbb{E}_{\bm{\epsilon}} \big[ ||\bm{\epsilon}_\theta(\bm{x}_{t(\lambda)}, t(\lambda)) - \bm{\epsilon}||_2^2 \big] ~\diffop \lambda.
\end{align}

While the reweighted variants still learn the correct score function that allows sampling, many of them no longer follow the original derivation of ELBO maximization for the data.
Kingma \etal~\cite{kingma2024understanding} shows certain conditions under which diffusion losses are equivalent to maximizing an ELBO objective with data augmentation:
\begin{align}
    \mathcal{L}(\bm{x}_0) = \mathbb{E}_{p_w(t)}[\mathcal{L}_t(\bm{x}_0)] + \textrm{constant}, \label{L_pw}
\end{align}
where $p_w(t)=\frac{\diffop}{\diffop t}w(\lambda_t)$ is a distribution, assuming $w(\lambda_t)$ is normalized such that $w(\lambda_1)=1$, and
\begin{align}
    \mathcal{L}_t(\bm{x}_0) &= D_{KL}(q(\bm{x}_{t\dots1}|\bm{x}_0)||p_\theta(\bm{x}_{t\dots1})) \\
    &\geq  D_{KL}(q(\bm{x}_t|\bm{x}_0)||p_\theta(\bm{x}_t)) \\
    &= -\mathbb{E}_{q(\bm{x}_t|\bm{x}_0)}[\log p_\theta(\bm{x}_t)] + \textrm{constant}. \label{eq:elbo}
\end{align}
The diffusion objective is ELBO maximization if $p_w(t)=\frac{\diffop}{\diffop t}w(\lambda_t)\geq 0$.

We base the theory of our diffusion tokenizers on the diffusion models with ELBO objectives, such as Flow Matching~\cite{lipman2023flow,albergo2022building,liu2022flow} as shown in Kingma \etal~\cite{kingma2024understanding}, which we detail in approach.

\section{Approach}
\label{sec:approach}

Our goal is to learn compressed latent representations of images that can be used for training latent-space image generation models.
This compression is learned via a tokenizer that can compress the image from pixel space to latent space (tokens) and decompress it from latent space to pixel space.
More formally, given an input image $\bm{x}$ in pixel space, it is passed into an encoder $E$ to obtain the compact latent representation or tokens $\bm{z}$. The latent $\bm{z}$ is used as the condition for a diffusion decoder $D$ that models the distribution $p(\bm{x}|\bm{z})$.
An overview of our diffusion tokenizer (\Ours) is shown in~\cref{fig:method}.

During training, a noisy image $\bm{x}_t$ is constructed by adding noise to $\bm{x}$ with the forward diffusion process at random time $t\in [0, 1]$, then the diffusion network $D$ takes both $\bm{x}_t$ and $\bm{z}$ as input and is supervised by the Flow Matching objective.
At test time, given a latent representation $\bm{z}$, the reconstruction image in pixel space can be decoded by first sampling Gaussian noise $\bm{\epsilon}\sim \mathcal{N}(\bm{0}, \bm{I})$, and then iteratively ``denoising'' it with reverse diffusion process conditioned on $\bm{z}$.
$E$ and $D$ are jointly trained from scratch to learn the latent representation and conditional decoding together.

\vspace{-0.5em}
\paragraph{Training objective.}
We follow Flow Matching~\cite{lipman2023flow,albergo2022building,liu2022flow} that is shown~\cite{kingma2024understanding} to be an ELBO maximization diffusion objective. The noise schedule is defined as
\begin{align}
    \alpha_t = 1 - t, \quad \sigma_t = \sigma_{\min} + t \cdot (1 - \sigma_{\min}),
\end{align}
where $\sigma_{\min}=10^{-5}$. The diffusion network $D$ uses $\bm{v}$-prediction~\cite{salimans2022progressive,lipman2023flow} that is trained with the objective
\begin{align}
    \label{eq:fm_obj}
    \mathcal{L}(\bm{x}) = \mathbb{E}_{t,\bm{\epsilon}} \big[ ||D(\bm{x}_t, t, \bm{z}) - \big((1-\sigma_{\min}) \bm{\epsilon} -\bm{x}\big)||_2^2 \big].
\end{align}
The time $t$ is uniformly sampled in $[0, 1]$.

\paragraph{Simple implementation.}
Our implementation only uses a single L2 loss (\cref{eq:fm_obj}).
Thus, unlike \gae, it does not require access to pretrained discriminative models to compute LPIPS loss, or training an extra GAN discriminator in an adversarial game.
Since we use a single loss, our method does not need a combinatorial search for loss weight rebalancing in contrast to \gae.
We also observe that discarding the variational KL regularization loss for $\bm{z}$ in \gae has negligible impact on \Ours.
Finally, \Ours is a self-supervised technique, unlike \gae that relies on pretrained supervised discriminative models in LPIPS.

\paragraph{Theoretical justification.}
A scalable autoencoder typically requires a principled objective.
We connect the finding from Kingma \etal~\cite{kingma2024understanding} to our diffusion autoencoder to show its theoretical basis.
Given the recent results~\cite{kingma2024understanding}, our choice of the Flow Matching training objective can be interpreted as learning to compress the image $\bm{x}$ into a latent $\bm{z}$ while maximizing the ELBO $\mathbb{E}_{q(\bm{x}_t|\bm{x})}[\log p_D(\bm{x}_t|\bm{z})]$.
That is, $\bm{z}$ is learned to maximize the log probability density of the input $\bm{x}$ augmented at all noise levels $t$ in the expectation.
The widely used $\bm{\epsilon}$-prediction (with cosine schedule)~\cite{nichol2021improved} and EDM~\cite{karras2022elucidating} are shown~\cite{kingma2024understanding} not in this ELBO form and may not directly maximize the log probability density of the input.
We study the effects of these objectives in our experiments and observe the practical benefits of the ELBO objectives.

\paragraph{Noise synchronization.}
We propose an additional regularization on the \Ours's latent representations $\bm{z}$ that facilitates training the latent diffusion model on top of them for image generation.
When these latents $\bm{z}$ are used to train the latent diffusion model, they are noised as $\bm{z}_t$.
While clean variables $\bm{z}_0$ are supervised to contain rich information for reconstruction by the diffusion decoder, the noising process from $t=0$ to $1$ on $\bm{z}_t$ may potentially destroy the information too quickly or slowly in an uncontrolled way.

To make the diffusion path for the latent variable $\bm{z}$ more smooth, we try to synchronize the noising process on the latent $\bm{z}$ to the pixel space $\bm{x}$.
The idea is to encourage the noisy $\bm{z}_\tau$ to maximize the ELBO for the noise images $\bm{x}_{\tau\dots 1}$ (\cref{eq:elbo}).
Specifically, during the \Ours training, after obtaining $\bm{z}=E(\bm{x})$, we augment $\bm{z}_\tau=\alpha_\tau \bm{z} + \sigma_\tau \bm{\epsilon}$ with probability $p=0.1$ for a random time $\tau\in[0, 1]$, then use the diffusion decoder to compute the denoising loss with $t$ sampled in $[\tau, 1]$. Intuitively, it encourages $\bm{z}_\tau$ to help denoising $\{\bm{x}_t\mid t\in [\tau, 1]\}$, where larger $\tau$ corresponds to denoising at higher noise levels, which are for more global and lower-frequency information.

\subsection{Implementation Details}
We describe the architecture and training hyperparameters for our diffusion tokenizers.

\vspace{-0.5em}
\paragraph{Architecture.} The encoder $E$ follows the standard convolutional encoder used in Stable Diffusion (LDM~\cite{rombach2022high}) and SDXL~\cite{podell2024sdxl}, with the configuration that has a spatial downsampling factor 8, and 4 channels for the latent. The decoder $D$ is a convolutional UNet with timestep conditioning that follows Consistency Decoder~\cite{yang2023consistency}. The $\bm{z}$-condition of the diffusion model is implemented by nearest upsampling $\bm{z}$ and concatenation to $\bm{x}_t$ as the input to the decoder. While the original autoencoder in LDM~\cite{rombach2022high} applies a KL loss on the latent as in a variational autoencoder, we remove it and simply use a LayerNorm~\cite{ba2016layernormalization} on $\bm{z}$, which eliminates the burden to balance an additional KL loss (see \cref{app:layernorm}).

\paragraph{Training.} Both the encoder and diffusion decoder are jointly trained from scratch. We use AdamW~\cite{loshchilov2018decoupled} optimizer, with constant learning rate $0.0001$, $\beta_1=0.9$, $\beta_2=0.999$, weight decay $0.01$. By default, diffusion tokenizers are trained for 300K iterations with batch size 64. We refer to more details in \cref{app:training}.

\paragraph{Inference.} We choose the Euler ODE solver for simplicity, and use 50 steps to sample from the diffusion decoder $D$.

\begin{table}
    \centering
    \begin{tabular}{c|lc}
        & \bf Model & \bf rFID@5K \\
        \hline
        \multirow{6}{*}{Supervised} & \cite{rombach2022high} & 4.37 \\
        & \gae-B & 4.39 \\
        & \gae-L & 4.05 \\
        & \gae-XL & 4.14 \\
        & \daeB \footnotesize{(+LPIPS)} & 4.13 \\
        & \daeXL \footnotesize{(+LPIPS)} & \textbf{3.53} \\
        \hline
        \multirow{3}{*}{Self-supervised} & \daeB & 8.91 \\
        & \daeL & 8.75 \\
        & \daeXL & 7.95 \\
    \end{tabular}
    \caption{\textbf{Comparison for image reconstruction on ImageNet.} While \Ours-XL shows a higher FID metric, it achieves better visual quality than \gae-XL (Fig.~\ref{fig:teaser},\ref{fig:human_eval}). When adding the supervised LPIPS loss (already used in \gae) to explicitly match deep network features, \Ours's FID outperforms \gae.}
    \label{tab:cmp_recon}
\end{table}

\begin{figure*}
    \centering
    \includegraphics[width=\linewidth]{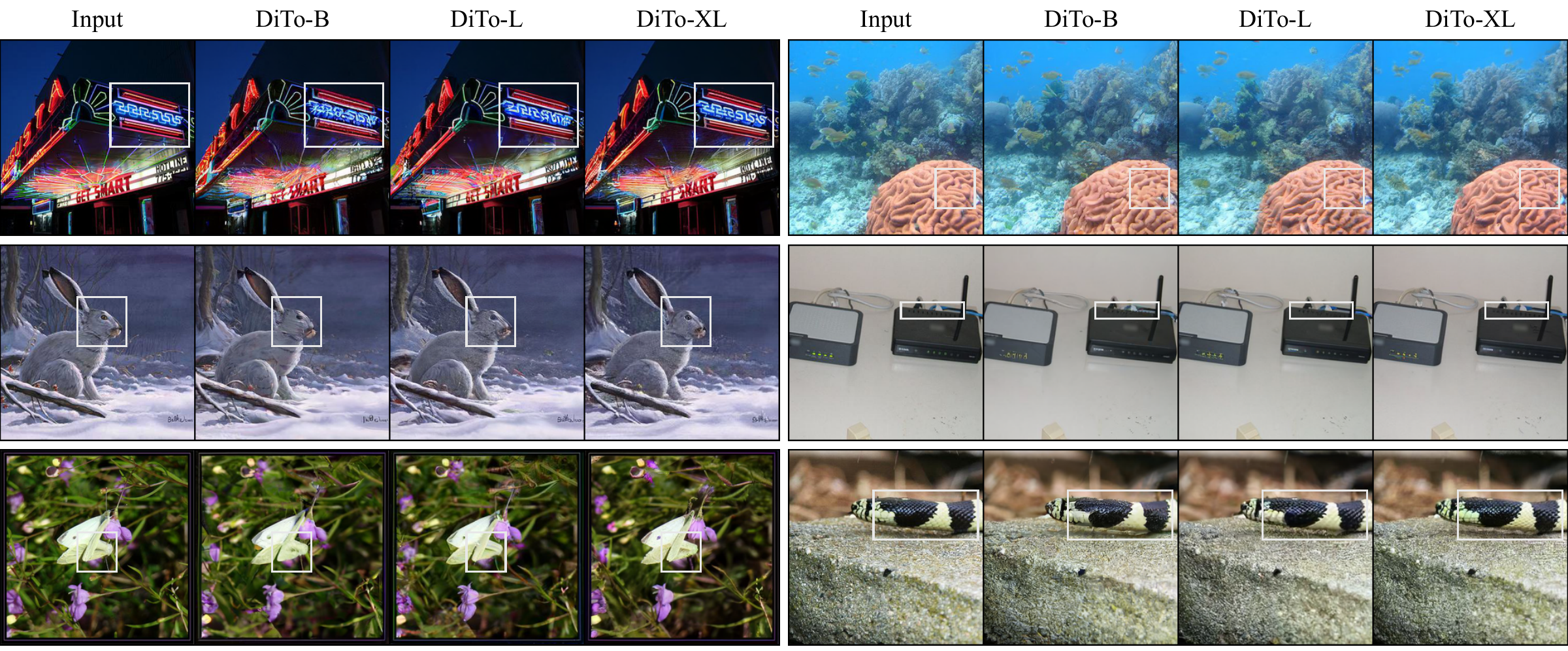}
    \caption{\textbf{Scalability of diffusion tokenizers.} When increasing the number of trainable parameters in the diffusion decoder from \daeB, \daeL, to \daeXL in the joint training, we observe that the image reconstruction quality keeps improving for structures and textures. Both the visual quality and reconstruction faithfulness are improved when scaling up the diffusion tokenizer.}
    \label{fig:scale}
\end{figure*}

\section{Experiments}
\label{sec:experiments}

\paragraph{Dataset.} We use the \imnet~\cite{deng2009imagenet} dataset, which is large-scale and contains diverse real-world images, to train and evaluate our models and baselines for both image reconstruction and generation.
We post-process the dataset such that faces in the images are blurred.
By default, images are resized to be at 256 pixel resolution for the shorter side. For tokenizer training, we apply random crop and horizontal flip as data augmentation. Images are center-cropped for evaluation.

\paragraph{Baselines.} We compare to the standard tokenizer used in LDM~\cite{rombach2022high}, which we refer to as \gae. It is widely used in recent state-of-the-art visual generative models~\cite{peebles2023scalable,karras2024analyzing,rombach2022high,podell2024sdxl}.
The tokenizer uses L1, LPIPS, and GAN loss for reconstruction. 
For a fair comparison, we train \gae using the same training data and the same architecture that matches the number of parameters to the corresponding \Ours model (see \cref{app:training}).
The \gae downsamples by a factor of $8$ and produces a latent $\bm{z}$ of size $4\times32\times32$.

\paragraph{Models.} Since the main difference of \Ours compared to the baselines is the diffusion decoder, we fix the encoder as the encoder in LDM~\cite{rombach2022high} with a downsampling factor 8 by default, and evaluate several variants of the diffusion decoder in different sizes, the settings are denoted as \daeB, \daeL, and \daeXL with 162.8M, 338.5M, 620.9M parameters in the decoder respectively. The architecture details are provided in~\cref{app:architecture}.
The same as \gae, \Ours's $\bm{z}$ is of the size $4\times32\times32$.

\paragraph{Automatic evaluation metrics.} We evaluate the commonly used Fréchet Inception Distance (FID)~\cite{heusel2017gans} for both the reconstruction and generation. The reconstruction FID (rFID) is computed between a set of input images and their corresponding reconstructed images by the tokenizer. The generation FID (gFID) is computed between randomly generated images and the dataset images. For computation efficiency, we use a fixed set of 5K images from \imnet validation set to evaluate rFID (which we observe to be stable, while it is typically higher than FID with 50K samples, see \cref{app:fid_50k}). We evaluate gFID with 50K samples.

\vspace{-1em}
\paragraph{Human evaluation.}
Recent work shows that automated metrics for evaluating visual generation do not correlate well with human judgment~\cite{girdhar2023emu,podell2024sdxl,borji2019pros,borji2022pros,jayasumana2024rethinking}.
Thus, we also collect human preferences to compare our method and baselines.
To compare the two models, we set up a side-by-side evaluation task where humans pick the preferred result.
We provide the details in~\cref{app:human_eval}.

\subsection{Image reconstruction}
\label{sec:experiments_reconstruction}

We compare the reconstruction quality of \Ours and the baseline \gae.
Reconstruction quality directly measures the ability of the tokenizer to learn compact latent representations (tokens) that can reconstruct the image.
\Ours is trained without noise synchronization (\cref{sec:approach}) by default as we measure the reconstruction quality in this section.

The qualitative results are shown in~\cref{fig:teaser}.
Despite using a simpler loss, we observe that \Ours shows a better reconstruction quality than \gae, especially for regular visual structures, symbols, and text, as shown in the example images.
A potential reason might be that the \gae relies on the heuristic LPIPS loss that matches the deep network features of the reconstructed image. While it is good for random textures, it may be not accurate enough for structured details.
\Ours has principled probabilistic modeling (ELBO) for decoding images, and thus can learn to compress the common patterns, including visual structures and text appearance by compressing images using the self-supervised reconstruction loss.

\begin{figure}
    \centering
    \includegraphics[width=.6\linewidth]{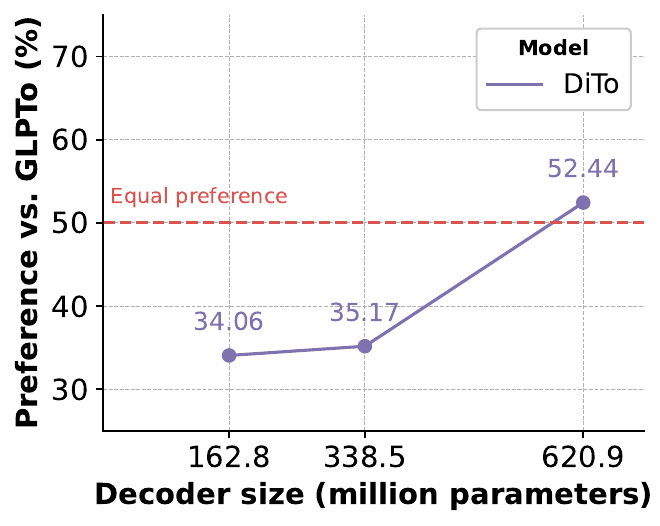}
    \caption{\textbf{Comparison for human preference of image reconstructions.} Models are compared to \gae at the same scale. When being scaled up, we observe that \Ours's (without perceptual loss) visual quality significantly improves and outperforms \gae in human preference.}
    \label{fig:human_eval}
\end{figure}

\begin{figure*}
    \centering
    \includegraphics[width=\linewidth]{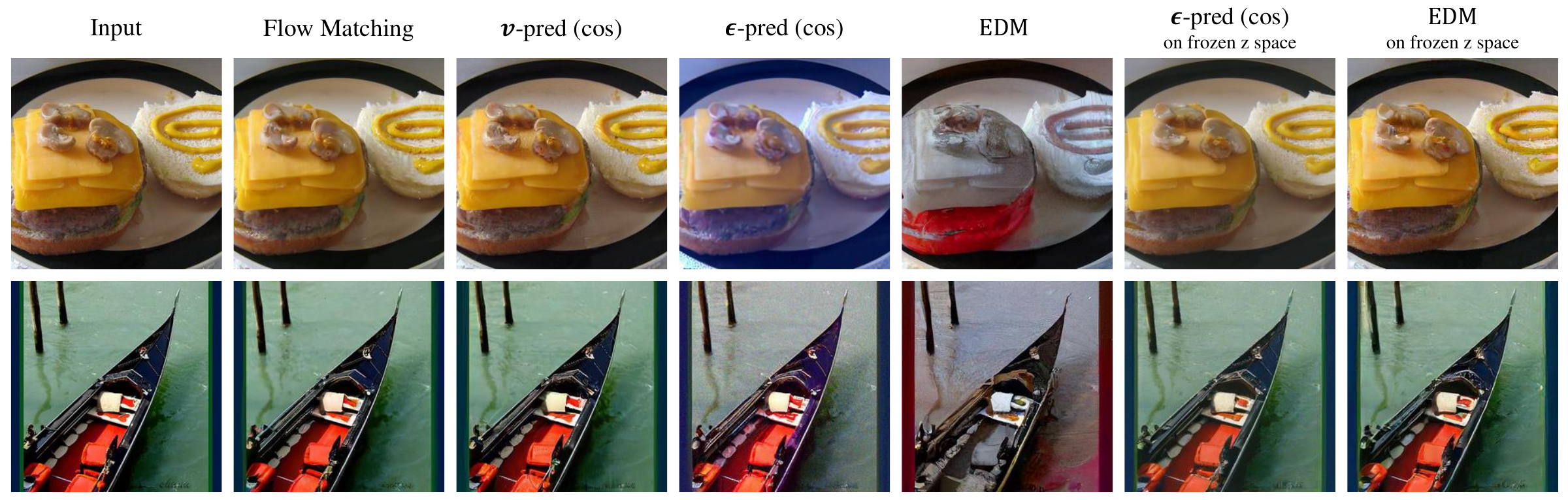}
    \caption{\textbf{Comparison of training objectives in diffusion tokenizers.} The frozen $\bm{z}$ space is from a \gae-B. We observe that when jointly training the encoder and diffusion decoder, ELBO diffusion objectives (flow matching, $\bm{v}$-pred with cosine schedule) can learn good latent representation $\bm{z}$, while other objectives may have color shift in the reconstruction (colors are good given a frozen $\bm{z}$ space).}
    \label{fig:objective}
\end{figure*}

A quantitative comparison is shown in~\cref{tab:cmp_recon}.
\Ours has a higher reconstruction FID than the \gae.
FID is computed using distance in a supervised deep network feature space.
We hypothesize that the LPIPS loss heuristic plays an important role in the \gae to achieve a low FID as it explicitly matches supervised deep network features for the reconstruction and the ground truth.
Based on this hypothesis, we train a variant of \Ours that uses an additional LPIPS loss (see \cref{app:dito_lpips}).
Note that LPIPS loss is typically necessary for stability and visual quality in \gae training, while it is optional for \Ours.
We observe that the supervised variant of \Ours with LPIPS loss achieves lowest FID while controlling for model size, \ie, \daeB with LPIPS outperforms a similarly sized \gae-B and \daeXL with LPIPS also outperforms \gae-XL.

\begin{table}[!t]
    \centering
    \begin{tabular}{lc|>{\color[gray]{0.8}}c}
        \multirow{2}{*}{\bf Latent encoder} & \multirow{2}{*}{\bf gFID@50K} & rFID@5K \\
        & & \small{(Autoencoding)} \\
        \hline
        \gae-XL & 7.49 & 4.14 \\
        \daeXL & 7.57 & 7.95 \\
        \daeXL \footnotesize{(w/ noise sync.)} & \textbf{6.29} & 8.65
    \end{tabular}
    \caption{\textbf{Training image generation models} on the latent representations from \Ours and \gae. We train DiT models and compare the image generations. We observe that the latent representations from \Ours lead to competitive image generations.
    Our proposed noise synchronization further improves the generation quality and outperforms the generations using a \gae.}
    \label{tab:cmp_gen}
\end{table}

\vspace{-1em}
\paragraph{Scalability.}
We study the scalability of \Ours on the three variants - \daeB, \daeL, and \daeXL, where we nearly double the decoder size across each model while keeping the encoder architecture unchanged. A qualitative comparison of the image reconstructions by these models is shown in~\cref{fig:scale}.
We observe that both the image reconstruction quality and the reconstruction faithfulness keep improving as the model is scaled up.
The improvements of scaling are also confirmed by the reduction in reconstruction FID in~\cref{tab:cmp_recon}, where the rFID smoothly reduces with model size.
However, as shown in~\cref{tab:cmp_recon}, FID is affected by the supervised LPIPS loss, and many recent works report that it is not aligned with visual quality~\cite{borji2019pros,borji2022pros,jayasumana2024rethinking}.
Thus, we use human evaluations to compare the self-supervised \Ours and the supervised \gae.

We conduct a side-by-side human evaluation of the image reconstructions from these models and report the preference rate in~\cref{fig:human_eval}, where a preference greater than $50\%$ indicates that a model `wins' over the other.
At sizes of B ($162.8$M) and L ($338.5$M), the supervised \gae's image reconstructions are preferred over those of \Ours.
However, when further scaling up to XL ($620.9$M), we observe that self-supervised \daeXL's reconstructions are preferred over the \gae-XL.
Qualitatively, we observed that the quality of \gae gets mostly saturated when scaling up the decoder and the failure cases are not significantly improved.
In contrast, we observed many reconstruction details keep improving for \Ours with the decoder size.
This result also shows that \Ours is a scalable, simpler, and self-supervised alternative to \gae.

Finally, we note that while evaluating reconstructions is meaningful, in the next step, the representations from \Ours and \gae are used to train image generation models.
We evaluate how useful these representations are for image generation in~\cref{sec:experiments_generation}.

\subsection{Image generation}
\label{sec:experiments_generation}

We compare the performance of training a latent diffusion image generation model on the learned latent representation $\bm{z}$ from either \Ours or \gae.
We follow DiT~\cite{peebles2023scalable} and use DiT-XL/2 as the latent diffusion model for class-conditioned image generation on the \imnet dataset (see more details in \cref{app:ldm_training}).
We compare the image generations from the resulting DiT models in~\cref{tab:cmp_gen} and draw several observations.

A DiT trained using \Ours without noise synchronization achieves competitive FID to a DiT trained using \gae suggesting that the latent image representations of \Ours are suitable for downstream image generation tasks.
Note that when compared in~\cref{tab:cmp_recon}, \Ours has a higher reconstruction FID than \gae with a larger gap.
It suggests that the low FID advantage achieved by explicitly matching deep features may not be fully inherited in the image generation stage. 
A DiT trained on \Ours with noise synchronization achieves the best performance, even outperforming \gae in FID.
This result confirms the effectiveness of \Ours as a tokenizer for image generation.

\begin{figure}
    \centering
    \includegraphics[width=\linewidth]{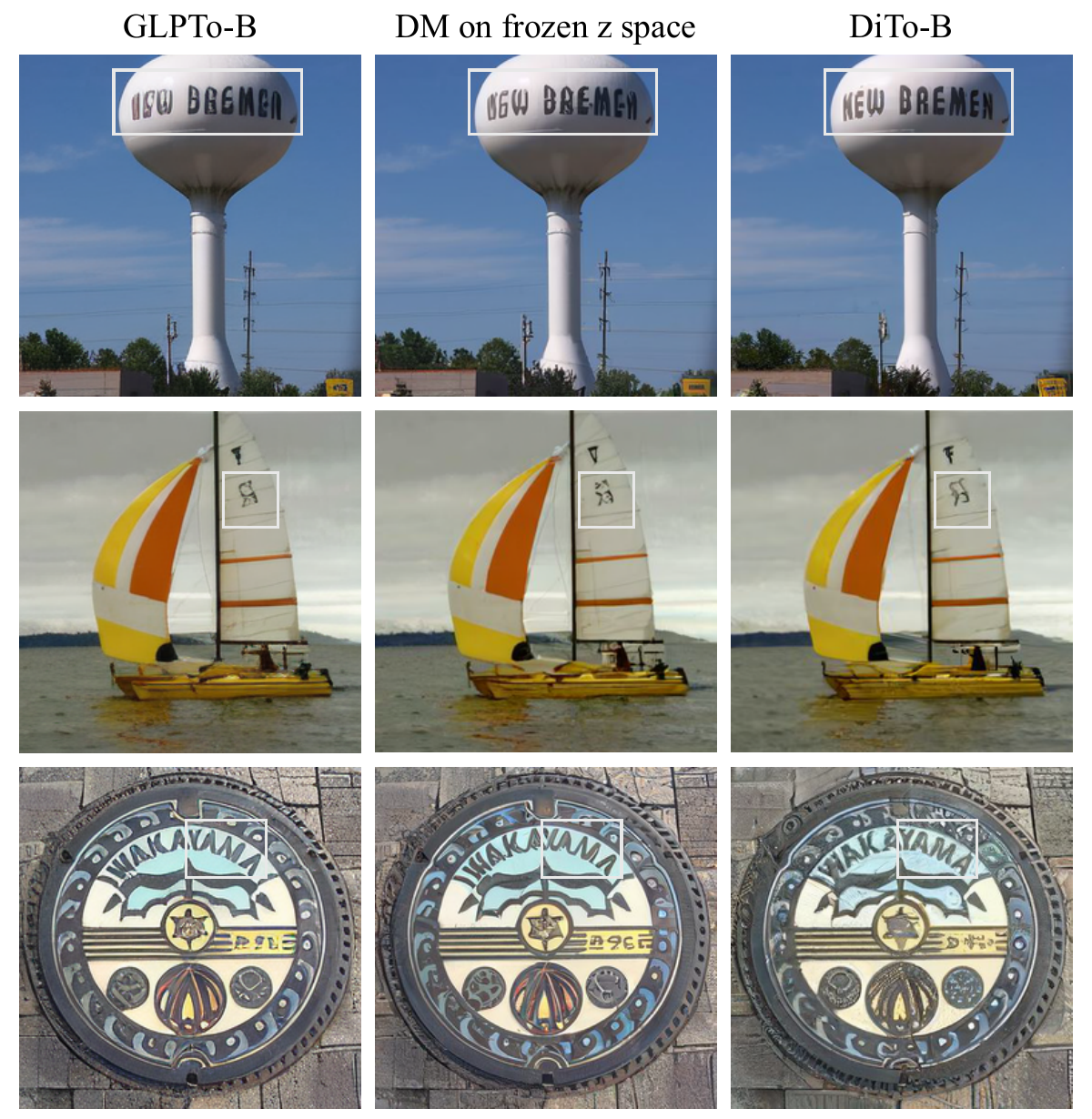}
    \caption{\textbf{Effectiveness of the latent representation \vs decoder.} 
    We train a \Ours decoder-only on a frozen latent space from \gae and observe that the reconstruction results are more similar to using a \gae decoder (notice similar errors on the visual text reconstruction). These reconstructions are qualitatively different compared to an end-to-end trained \Ours's reconstructions.
    This suggests that the effectiveness of \Ours comes from jointly learning a powerful decoder and a latent representation.}
    \label{fig:latent}
    \vspace{-1em}
\end{figure}

\subsection{Ablations and Analysis}
\label{sec:ablations}

We now present ablations of our design choices and analyze the key components of \Ours.
We follow the same experimental setup as in~\cref{sec:experiments_reconstruction}.

\vspace{-1em}
\paragraph{Training objectives.}
As described in~\cref{sec:preliminaries,sec:approach}, our \Ours uses a Flow Matching objective which can be viewed as an ELBO maximization for image reconstruction.
In contrast, as shown in~\cite{kingma2024understanding}, the widely used diffusion implementations such as $\bm{\epsilon}$-prediction (with cosine noise schedule) and EDM~\cite{karras2022elucidating} are not ELBO objectives.
We now study the impact of this by training three variants of \Ours and changing the training objective only.
We show the examples of the reconstructions in~\cref{fig:objective}.
Using the ELBO objectives of Flow Matching and $\bm{v}$-prediction (with cosine schedule, which is also an ELBO objective) yields image reconstructions that are more faithful to the input image.
The non-ELBO objectives of $\bm{\epsilon}$-prediction and EDM yield reconstructions sometimes with a noticeable loss of faithfulness, \eg, color shift.
To further investigate this, we start with a pretrained \gae encoder and keep it frozen while learning diffusion decoders from scratch with the different training objectives.
We observe that the image reconstructions do not have such obvious color shift, suggesting that the non-ELBO objectives can `decode' correctly but may lead to learning sub-optimal latent representations.
A potential reason might be that the non-ELBO objectives have a non-monotonic weight function $w(\lambda)$ for different log SNR ratios, which makes some terms contribute negatively in~\cref{L_pw}, and leads to training noise or bias for reconstruction.

\vspace{-1em}
\paragraph{Effectiveness of the latent representation \vs decoder.}
We now study whether the effectiveness of \Ours \vs \gae mainly comes from the decoder's powerful probabilistic modeling or from jointly learning both a powerful latent $\bm{z}$ and the decoder.
We train a \Ours decoder-only on a frozen latent space from a \gae and compare the reconstructions to the \gae in~\cref{fig:latent}.
We observe that both reconstructions look qualitatively similar, and have the same error modes around visual text reconstruction.
When compared with reconstructions from an end-to-end \Ours, we observe qualitative differences, \eg, the visual text reconstruction is clearer.
This suggests that \Ours's effectiveness lies in jointly learning a powerful latent $\bm{z}$ that is helpful to the probabilistic reconstruction objective of the decoder.

\section{Conclusion and Discussion}

We showed that diffusion autoencoders with proper design choices can be scalable tokenizers for images. Our diffusion tokenizer (\Ours) is simple, and theoretically justified compared to prior state-of-the-art \gae.
\Ours training is self-supervised compared to the supervised training (LPIPS) from \gae.
Compared to \gae, we observe that \Ours's learned latent representations achieve better image reconstruction, and enable better downstream image generation models.
We also observed that \Ours is easier to scale and its performance improves significantly with scale.

There are several directions to be further explored for diffusion tokenizers.
Our work only explored learning tokenizers for a downstream image generation task.
We believe learning tokenizers that work well for both recognition and generation tasks will greatly simplify model training.
We also believe content-aware tokenizers that can encode the spatially variable information density in images will likely lead to higher compression.
Finally, this paper only studies diffusion tokenizers for images. We believe extending this concept to video, audio, and other continuous signals will unify and simplify training.

\vspace{-0.5em}
\section*{Social Impact}

Our method is developed for research purpose, any real world usage requires considering more aspects. \Ours is an image tokenizer, the reconstructed image is perceptually similar but not exactly the same as the input image. The generative diffusion decoder and latent diffusion model may learn unintentional bias present in the dataset statistics.

\bibliography{main}
\bibliographystyle{icml2025}

\newpage
\appendix
\onecolumn

\section{Experiment Details}
\subsection{Architecture}
\label{app:architecture}

\begin{table}
    \centering
    \begin{tabular}{c|ccccc}
        \textbf{Model} & \textbf{\#Params} & $c_1$ & $c_2$ & $c_3$ & $t_{\textrm{emb}}$ \\
        \hline
        DiTo-B & 162.8M & 128 & 256 & 512 & 1280 \\
        DiTo-L & 338.5M & 192 & 384 & 768 & 1280 \\
        DiTo-XL & 620.9M & 320 & 640 & 1024 & 1280
    \end{tabular}
    \caption{Configuration details of the UNet diffusion decoder in \Ours at different scales.}
    \label{tab:arch_details}
\end{table}

\begin{figure}
    \centering
    \includegraphics[width=0.5\linewidth]{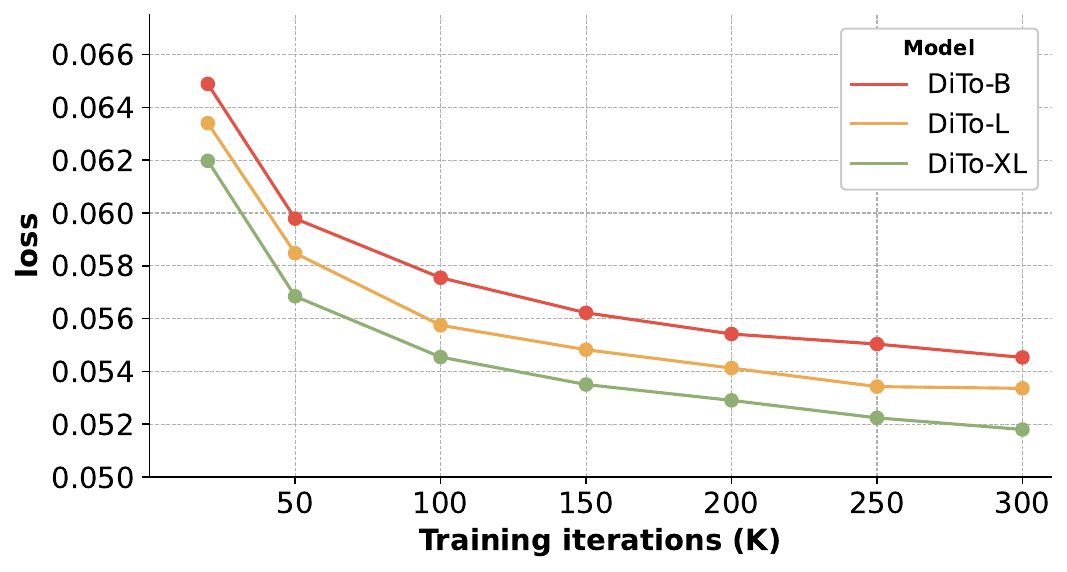}
    \caption{\textbf{Training loss curves of \Ours at different scales.} We observe the loss keeps improving as scaling up the model and the improvement is not saturated yet. The objective is Flow Matching and the loss is averaged over the latest 10K iterations.}
    \label{fig:train_loss}
\end{figure}

We follow the encoder in LDM~\cite{rombach2022high} and the decoder in Consistency Decoder~\cite{yang2023consistency}. Both the encoder and decoder are fully convolutional. The UNet diffusion network contains 4 stages, each stage contains 3 residual blocks. In the downsampling phase of the UNet, stages 1 to 3 are followed by an additional residual block with downsampling rate 2. The number of channels in 4 stages are $c_1, c_2, c_3, c_3$ correspondingly. The upsampling phase of the UNet is in reverse order accordingly. The time in the diffusion process is projected to a vector with $t_{\textrm{emb}}$ dimension and modulates the convolutional residual blocks. The configurations used for our diffusion tokenizers are summarized in \cref{tab:arch_details}.

\subsection{Tokenizer training}
\label{app:training}

In the tokenizer training stage, the model is trained with batch size 64 for 300K iterations, which takes about 432, 864, 1728 NVIDIA A100 hours for \Ours-B, \Ours-L, and \Ours-XL models correspondingly. The training loss curves are shown in \cref{fig:train_loss}. When scaling up the model, the loss of flow matching objective keeps improving and we did not observe it to be saturated yet. The corresponding baselines \gae-B, \gae-L, and \gae-XL take longer time per training iteration than their \Ours counterparts due to their additional LPIPS and discriminator networks. For all \gae, we follow the standard training setting in LDM~\cite{rombach2022high} for models with downsampling factor 8, where the loss weights $\lambda_{\textrm{L1}}=1.0$, $\lambda_{\textrm{LPIPS}}=1.0$, $\lambda_{\textrm{GAN}}=0.5$, the gradient norm of regression loss (L1 + LPIPS) and GAN loss are adaptively balanced during training, and the GAN loss is enabled after 50K iterations. To speed up training, we use mix-precision training with bfloat16.

\subsection{Latent diffusion model training}
\label{app:ldm_training}

We train DiT-XL/2~\cite{peebles2023scalable} as the latent diffusion model for the learned latent space. We follow the standard setting that uses batch size 256, Adam optimizer with learning rate $1\cdot 10^{-4}$, no weight decay, and horizontal flips as data augmentation. Flow Matching is used as the training objective. We use classifier-free guidance 2 to generate the samples. To efficiently compare the models, the latent diffusion model is trained for 400K iterations for all tokenizers.

\subsection{Human evaluation}
\label{app:human_eval}

We use Amazon Mechanical Turk (MTurk) to collect human preferences for reconstruction and compare the methods. In the evaluation interface, we first present a few examples of better reconstructions and equally good reconstructions, where for better reconstructions, the number of examples is equal for different methods. The worker is presented with three images in a row, with tags ``input image'', ``reconstruction 1'', and ``reconstruction 2''. Two reconstructions are from two different methods and are randomly shuffled with 0.5 probability. The worker is asked to select the better reconstruction result based on: (i) the faithfulness of contents to the input image; and (ii) the visual quality of the reconstructed image. There are three available options on the interface: (i) reconstruction 1; (ii) reconstruction 2; and (iii) very hard to tell which is better.

For the comparison of each model pair, we collect 900 preference results. The results in detail are shown in \cref{tab:human_details}, where ``$>$'' means \Ours is preferred than \gae and ``$=$''  means equal preference (option (iii)). We count ``$\geq$'' as the number of ``$>$'' plus half of the number of ``$=$''. From the results, we observe that \Ours with LPIPS loss (which is used in \gae) is competitive to \gae at B size and outperforms \gae at larger XL size. \Ours significantly improves as scaling up and outperforms \gae at XL size.

\section{Ablation on LayerNorm}
\label{app:layernorm}

Unlike \gae~\cite{rombach2022high} which uses a KL regularization loss on the latent $z$, in \Ours we only apply LayerNorm on the latent representation $z$ for both tokenizer and latent diffusion model training. The ablation on this design choice is shown in \cref{tab:ablation_layernorm}. We observe that using LayerNorm has a better reconstruction performance than KL loss for \Ours, and has a competitive performance for image generation. While the weight of KL loss is originally optimized for \gae and further sweeping the weight for \Ours may potentially improve the result, we choose LayerNorm for simplicity. There are several main reasons for using LayerNorm in \Ours: (i) The KL loss introduces an additional loss weight to tune, which is not convenient in practice; (ii) Noise synchronization supervises $\bm{z}_t=\alpha_t \bm{z}_0 + \sigma_t \bm{\epsilon}$, LayerNorm ensures $\bm{z}_0$ to have 0 mean and 1 std so that it does not collapse to the trivial solution; (iii) LayerNorm shows a better reconstruction performance. Moreover, with LayerNorm, the latent representation $z$ no longer needs to be normalized for training the latent diffusion model.

\section{Comparison to rFID with 50K Samples}
\label{app:fid_50k}

For computation efficiency, we evaluate the reconstruction FID on a fixed set of 5K samples. In \cref{tab:fid_50k}, we compare the metric evaluated with 5K samples and 50K samples. The FID evaluated with 50K samples typically has a smaller value than the one evaluated with 5K samples, while it preserves the order in comparison between different methods. We observe FID with 5K samples to be stable enough to compare different checkpoints of the same method or different methods, therefore we mainly compare with FID@5K for more efficient evaluation.

\section{Evaluation on other metrics} We evaluate the autoencoder models on other common metrics for reconstruction, the results are shown in \cref{tab:eval_other_metrics}. Note that GLPTo-XL and DiTo-XL (+LPIPS) are trained with the LPIPS loss. We observe that DiTo-XL has the best PSNR and SSIM. For the metrics associated with the deep network features, LPIPS and Inception Score (IS), GLPTo-XL and DiTo-XL (+LPIPS) achieve better results as they explicitly match the deep features in training (LPIPS loss), while DiTo-XL (+LPIPS) achieves the best results on LPIPS and IS.

\begin{table}
    \small
    \centering
    \begin{tabular}{c|cc>{\color[gray]{0.8}}c|c>{\color[gray]{0.8}}c}
        \multirow{2}{*}{\textbf{Model}} & \multicolumn{5}{c}{\textbf{Preference vs. \gae (\%)}} \\
         \cline{2-6}
         & $=$ & $>$ & $<$ & $\geq$ & $\leq$ \\
         \hline
         \Ours-B \footnotesize{(+LPIPS)} & 26.22 & 34.33 & 39.44 & 47.44 & 52.56 \\
         \Ours-XL \footnotesize{(+LPIPS)} & 22.11 & 42.56 & 35.33 & 53.61 & 46.39 \\
         \hline
         \Ours-B & 27.22 & 20.44 & 52.33 & 34.06 & 65.94 \\
         \Ours-L & 22.56 & 23.89 & 53.56 & 35.17 & 64.83 \\
         \Ours-XL & 19.33 & 42.78 & 37.89 & 52.44 & 47.56
    \end{tabular}
    \caption{\textbf{Human evaluation results in detail.} Models are compared to the \gae at the corresponding size. ``$>$'' means \Ours is preferred than \gae, ``$=$'' means equal preference. ``$\geq$'' is counted as the value of ``$>$'' plus half of the value of ``$=$''.}
    \label{tab:human_details}
\end{table}

\begin{table}
    \centering
    \begin{tabular}{l|cc}
        \textbf{Model} & \textbf{rFID@5K} & \textbf{gFID@5K} \\
        \hline
        \Ours-B \footnotesize{(KL loss)} & 13.50 & 17.96 \\
        \Ours-B \footnotesize{(LayerNorm)} & 8.91 & 17.00
    \end{tabular}
    \caption{\textbf{Ablation on \Ours's latent space regularization.} rFID is evaluated for autoencoder reconstruction. gFID is evaluated for image generation.}
    \label{tab:ablation_layernorm}
\end{table}

\begin{table}
    \centering
    \begin{tabular}{l|cccc}
        \textbf{Model} & \textbf{PSNR} ($\uparrow$) & \textbf{SSIM} ($\uparrow$) & \textbf{LPIPS} ($\downarrow$) & \textbf{IS} ($\uparrow$) \\
        \hline
        GLPTo-XL & 24.82 & 0.7434 & 0.1528 & 127.06 \\
        DiTo-XL \footnotesize{(+LPIPS)} & 24.10 & 0.7061 & 0.1017 & 128.05 \\
        DiTo-XL & 25.92 & 0.7646 & 0.2304 & 109.13 \\
    \end{tabular}
    \caption{\textbf{Evaluation on other metrics for reconstruction.} Note that GLPTo-XL and DiTo-XL (+LPIPS) are trained with the LPIPS loss.}
    \label{tab:eval_other_metrics}
\end{table}

\begin{table}
    \centering
    \begin{tabular}{l|cc}
        \textbf{Model} & \textbf{rFID@5K} & \textbf{rFID@50K} \\
        \hline
         \gae-XL & 4.14 & 1.24 \\
         \Ours-XL \footnotesize{(+LPIPS)} & 3.53 & 0.78 \\
         \Ours-XL & 7.95 & 3.26
    \end{tabular}
    \caption{\textbf{Comparison to reconstruction FID (rFID) evaluated with 50K samples.} rFID@50K typically has a lower value than rFID@5K, while it is consistent with rFID@5K (for a fixed set) and preserves the order for comparison.}
    \label{tab:fid_50k}
\end{table}

\section{DiTo with LPIPS Loss}
\label{app:dito_lpips}

In \Ours, the diffusion decoder is trained with Flow Matching objective and does not directly output an image. To apply the LPIPS loss, we need to first convert it to the diffusion model's sample-prediction $\bar{\bm{x}}=\mathbb{E}_{q(\bm{x}_0,\bm{\epsilon},\bm{x}_t)}[\bm{x}_0|\bm{x}_t]$, then supervise the sample-prediction with LPIPS loss, so that the gradient can be backpropagated. In general scenarios of diffusion decoders, assume the diffusion network prediction $\bm{f}_\theta(\bm{x}_t)$ is minimizing the L2 loss to $A_t \bm{x}_0 + B_t \bm{\epsilon}$, we have
\begin{align}
    \begin{bmatrix}
        \bm{x}_t \\
        \bm{f}_{\theta^*}(\bm{x}_t)
    \end{bmatrix}
    =
    \begin{bmatrix}
        \alpha_t & \sigma_t \\
        A_t & B_t
    \end{bmatrix}
    \begin{bmatrix}
        \bar{\bm{x}} \\
        \bar{\bm{\epsilon}}
    \end{bmatrix},
    \label{eq:bars}
\end{align}
where $\bar{\bm{\epsilon}}=\mathbb{E}_{q(\bm{x}_0,\bm{\epsilon},\bm{x}_t)}[\bm{\epsilon}|\bm{x}_t]$, $\bm{f}_{\theta^*}$ is the network prediction at optimal network point $\theta^{*}$. This is because
\begin{align*}
    \bm{x}_t &= \mathbb{E}_{q(\bm{x}_0,\bm{\epsilon},\bm{x}_t)}[\bm{x}_t|\bm{x}_t] \\
    &= \mathbb{E}_{q(\bm{x}_0,\bm{\epsilon},\bm{x}_t)}[\alpha_t \bm{x}_0 + \sigma_t \bm{\epsilon} |\bm{x}_t] \\
    &= \alpha_t \cdot \mathbb{E}_{q(\bm{x}_0,\bm{\epsilon},\bm{x}_t)}[\bm{x}_0 |\bm{x}_t] + \sigma_t \cdot \mathbb{E}_{q(\bm{x}_0,\bm{\epsilon},\bm{x}_t)}[\bm{\epsilon} |\bm{x}_t] \\
    &= \alpha_t \bar{\bm{x}} + \sigma_t \bar{\bm{\epsilon}},
\end{align*}
and the optimal prediction under L2 loss is
\begin{align*}
    \bm{f}_{\theta^*}(\bm{x}_t) &= \mathbb{E}_{q(\bm{x}_0,\bm{\epsilon},\bm{x}_t)}[A_t \bm{x}_0 + B_t \bm{\epsilon}|\bm{x}_t] \\
    &= A_t \cdot \mathbb{E}_{q(\bm{x}_0,\bm{\epsilon},\bm{x}_t)}[\bm{x}_0|\bm{x}_t] + B_t \cdot \mathbb{E}_{q(\bm{x}_0,\bm{\epsilon},\bm{x}_t)}[\bm{\epsilon}|\bm{x}_t] \\
    &= A_t \bar{\bm{x}} + B_t \bar{\bm{\epsilon}}.
\end{align*}
According to \cref{eq:bars}, we have
\begin{align}
    \begin{bmatrix}
        \bar{\bm{x}} \\
        \bar{\bm{\epsilon}}
    \end{bmatrix}
    =
    \begin{bmatrix}
        \alpha_t & \sigma_t \\
        A_t & B_t
    \end{bmatrix}^{-1}
    \begin{bmatrix}
        \bm{x}_t \\
        \bm{f}_{\theta^*}(\bm{x}_t)
    \end{bmatrix},
\end{align}
In the Flow Matching we used,
\begin{align}
    \begin{bmatrix}
        \alpha_t & \sigma_t \\
        A_t & B_t
    \end{bmatrix}^{-1}
    =
    \begin{bmatrix}
        1 - t & t \\
        -1 & 1
    \end{bmatrix}^{-1}
    =
    \begin{bmatrix}
        1 & -t \\
        1 & 1-t
    \end{bmatrix}.
\end{align}
Therefore, the sample prediction is
\begin{align}
    \bar{\bm{x}}_\theta(\bm{x}_t) = \bm{x}_t - t \cdot f_{\theta}(\bm{x}_t),
\end{align}
on which we apply the LPIPS loss. Intuitively, it can be also viewed as a ``one-step generation'' under $\bm{v}$-prediction. Our weight for the LPIPS loss is 0.5.

\begin{figure*}
    \centering
    \includegraphics[width=\linewidth]{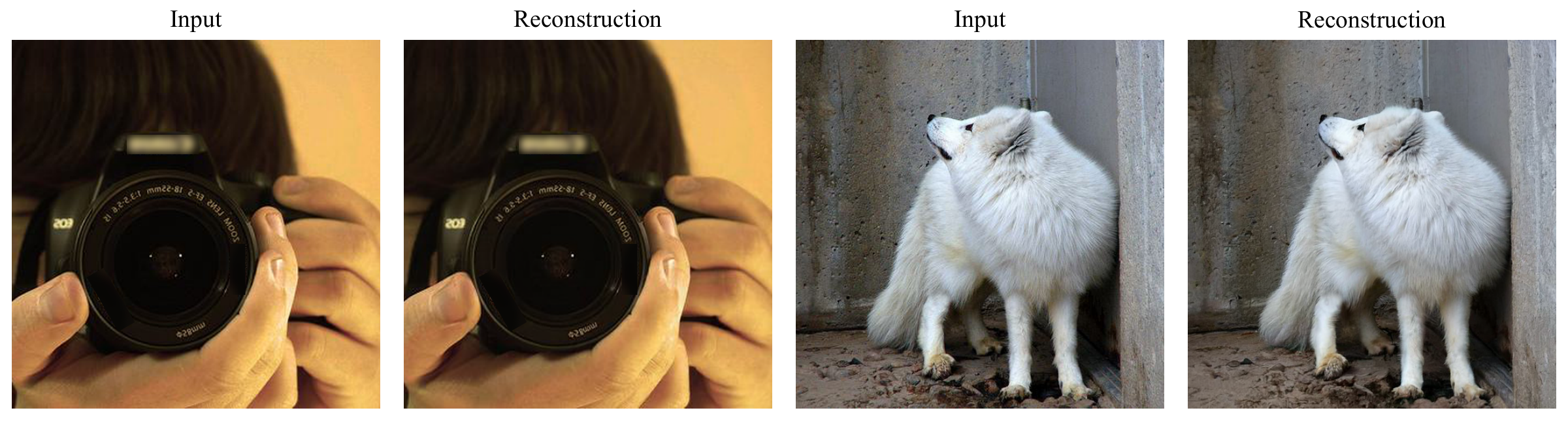}
    \caption{\textbf{Zero-shot generalization to tokenizing images at higher resolution.} Our diffusion tokenizer is fully convolutional and thus can generalize to autoencoding images at resolutions higher than the training setting (256 pixels) in zero-shot. The resolution is $512\times 512$ in the shown examples. A quantitative evaluation is shown in \cref{tab:recon_512}.}
    \label{fig:recon_512}
\end{figure*}

\section{Zero-Shot Generalization of Tokenization for Higher-Resolution Images}

We observe that our diffusion tokenizer, when trained for images at fixed $256\times 256$ pixels resolution, can generalize to a higher resolution at inference time (without any further training). We show examples for generating images at $512\times 512$ resolution in \cref{fig:recon_512}, and evaluate the corresponding reconstruction FID in \cref{tab:recon_512}. From the quantitative results at $512\times 512$ pixels resolution, we observe that the reconstruction FID gap between \Ours and \gae is significantly closed when generalizing to the higher resolution (from 7.95 vs. 4.14 to 2.32 vs. 2.13).

\begin{table}
    \centering
    \begin{tabular}{l|cc}
        \multirow{2}{*}{\textbf{Model}} & \multicolumn{2}{c}{\textbf{rFID@5K}} \\
         & $256\times 256$ & $512\times 512$ \\
        \hline
        \cite{rombach2022high} & 4.37 & 2.54 \\
        \gae-XL & 4.14 & 2.13 \\
        \Ours-XL & 7.95 & 2.32
    \end{tabular}
    \caption{\textbf{Quantitative comparison of zero-shot generalization to tokenizing images at higher resolution.} Similar to \gae, \Ours can also generalize to resolutions higher than training. We observe the rFID gap is significantly closed when evaluating at $512\times 512$ resolution.}
    \label{tab:recon_512}
\end{table}

\section{Motivation of Connecting to ELBO Theory}

We explain the motivation for connecting diffusion tokenizers to the recent ELBO theory~\cite{kingma2024understanding} in this section. In theory, given a fixed target distribution, general diffusion models with arbitrary weighting for log signal-to-noise ratios (SNR) can learn the correct score functions, which allow the model to sample from the target distribution. However, in the joint training of diffusion tokenizers, it is not clear what information is encouraged to be in $\bm{z}$ when the diffusion decoder is learning the score function of $p(\bm{x}_t|\bm{z})$. We note that in the view of learning score functions, it is not directly maximizing the probability $p(\bm{x}|\bm{z})$. Take $\bm{\epsilon}$-prediction as an example, it optimizes
\begin{align}
    \mathcal{L} = \mathbb{E}_{q(\bm{x}_0,\bm{\epsilon},\bm{x}_t)} \big[ ||\bm{\epsilon}_\theta(\bm{x}_t, t) - \bm{\epsilon}||_2^2 \big],
\end{align}
and ensures that
\begin{align}
    \bm{\epsilon}_{\theta^*}(\bm{x}_t, t)=\mathbb{E}_{q(\bm{x}_0,\bm{\epsilon},\bm{x}_t)}[\bm{\epsilon}|\bm{x}_t]
\end{align} at the optimal point $\theta^{*}$. The loss cannot and does not have to be zero, and a smaller loss does not necessarily mean more accurate score functions. Under specific conditions for effective log SNRs weighting (which depends on prediction type, noise schedule, and explicit time weighting), the loss becomes ELBO maximization objective, and minimizing the loss gets a theoretical guarantee.

Intuitively, while learning some representation $\bm{z}$ that is helpful to denoising $\bm{x}_t$ seems to be related to reconstructing $\bm{x}$, the weights for different times are crucial for learning $\bm{z}$. For example, if the weights at small times are too high, $\bm{z}$ may not need to store the global information of $\bm{x}$ (\eg, the overall color), as such information is always available at small noise levels while $\bm{z}$ only has limited capacity to be allocated. As a result, the reconstruction may have color shifts since $\bm{z}$ does not contain sufficient such information. Therefore, we propose to use the diffusion objectives that: (i) are shown to be stable and scalable in prior works; and (ii) are ELBO maximization objectives.

\begin{figure*}
    \centering
        {\includegraphics[width=0.29\linewidth]{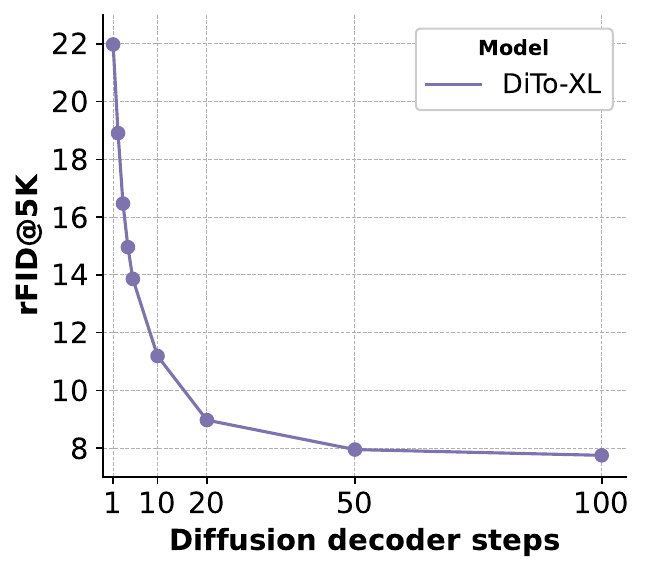}}
    \hfill
        {\includegraphics[width=0.6875\linewidth]{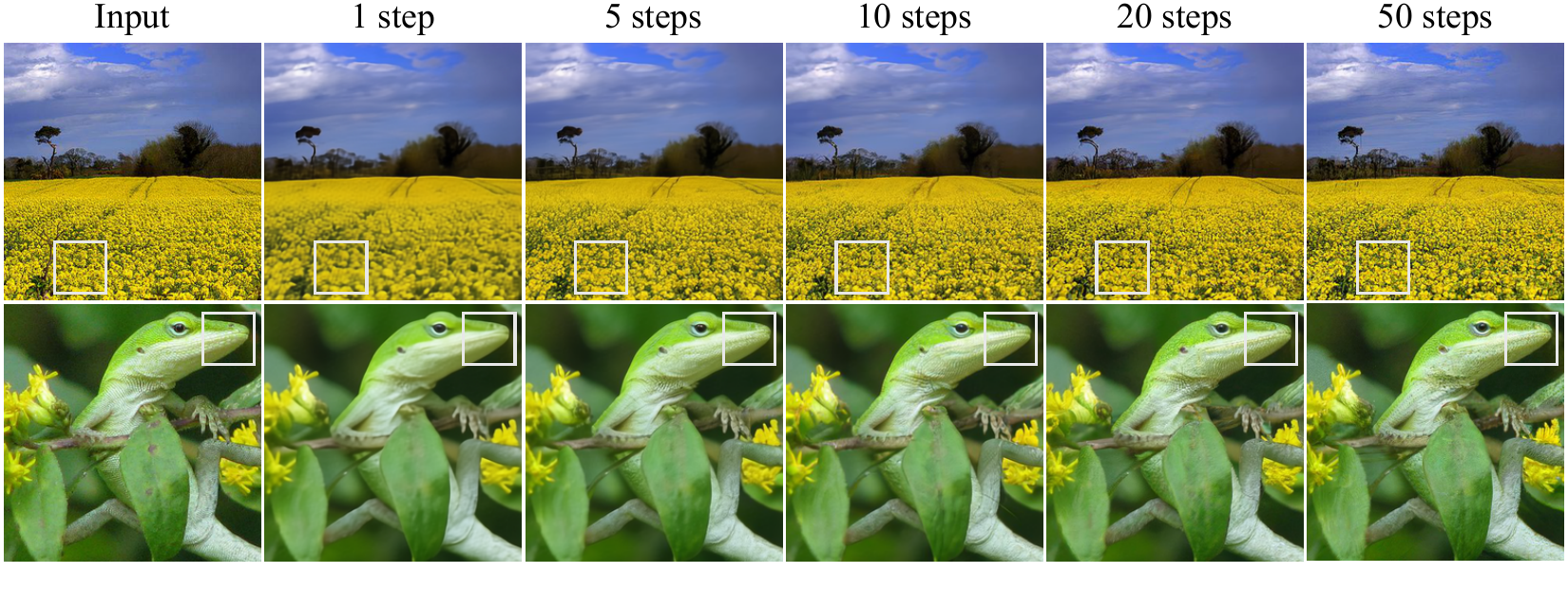}}
    \caption{\textbf{Number of decoding steps \vs image reconstruction quality.} We vary the number of steps in \Ours's diffusion decoder used for image reconstruction.
    We use the simple Euler ODE solver and observe that 20 to 50 steps are generally sufficient for good reconstruction quality. The rFID metric mostly converges after 50 steps, while the visual differences are not obvious after 10 to 20 steps.}
    \label{fig:decoding_steps}
\end{figure*}

\section{Number of decoding steps \vs quality of reconstruction.}
The decoder in \Ours is a diffusion model that reconstruct the image from the latent $\bm{z}$ using an iterative denoising process.
We study the effect of the number of iterations, \ie, decoding steps on the image reconstruction quality in~\cref{fig:decoding_steps}.
As expected, the image reconstruction quality improves with more steps (indicated by a lower rFID).
However, the visual quality mostly saturates after about 20 steps.
Since the iterative process can be computationally expensive, one-step diffusion distillation methods~\cite{song2024improved,yin2024onestep,xie2024distillation,salimans2024multistep,yin2024improved,lu2024simplifying} may be applied to speed up decoding in the future.

\section{Additional Qualitative Results}

\begin{figure*}
    \centering
    \includegraphics[width=\linewidth]{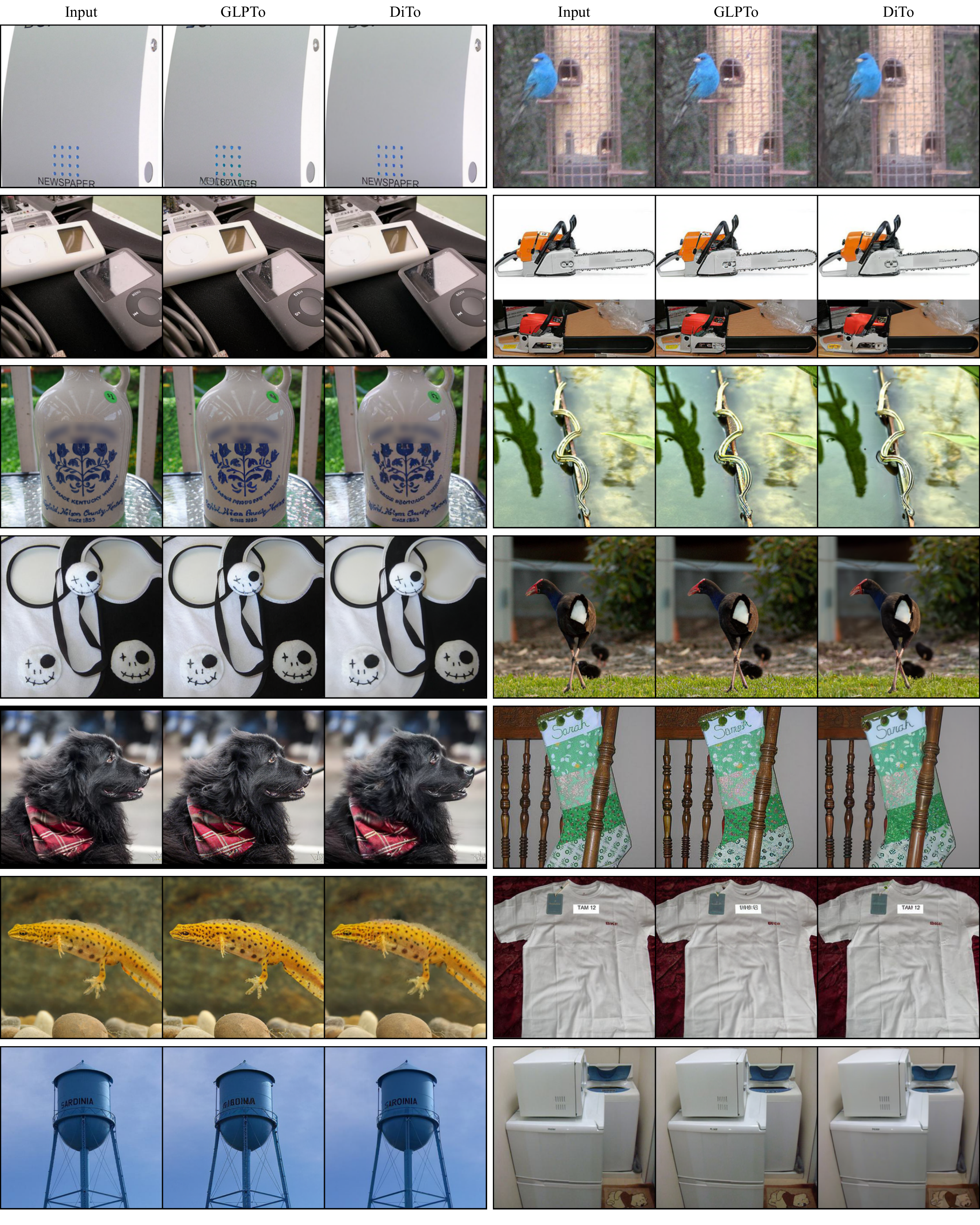}
    \caption{Additional qualitative comparison of tokenizers (at 256 pixel resolution).}
    \label{fig:supp_cmp_1}
\end{figure*}

\begin{figure*}
    \centering
    \includegraphics[width=\linewidth]{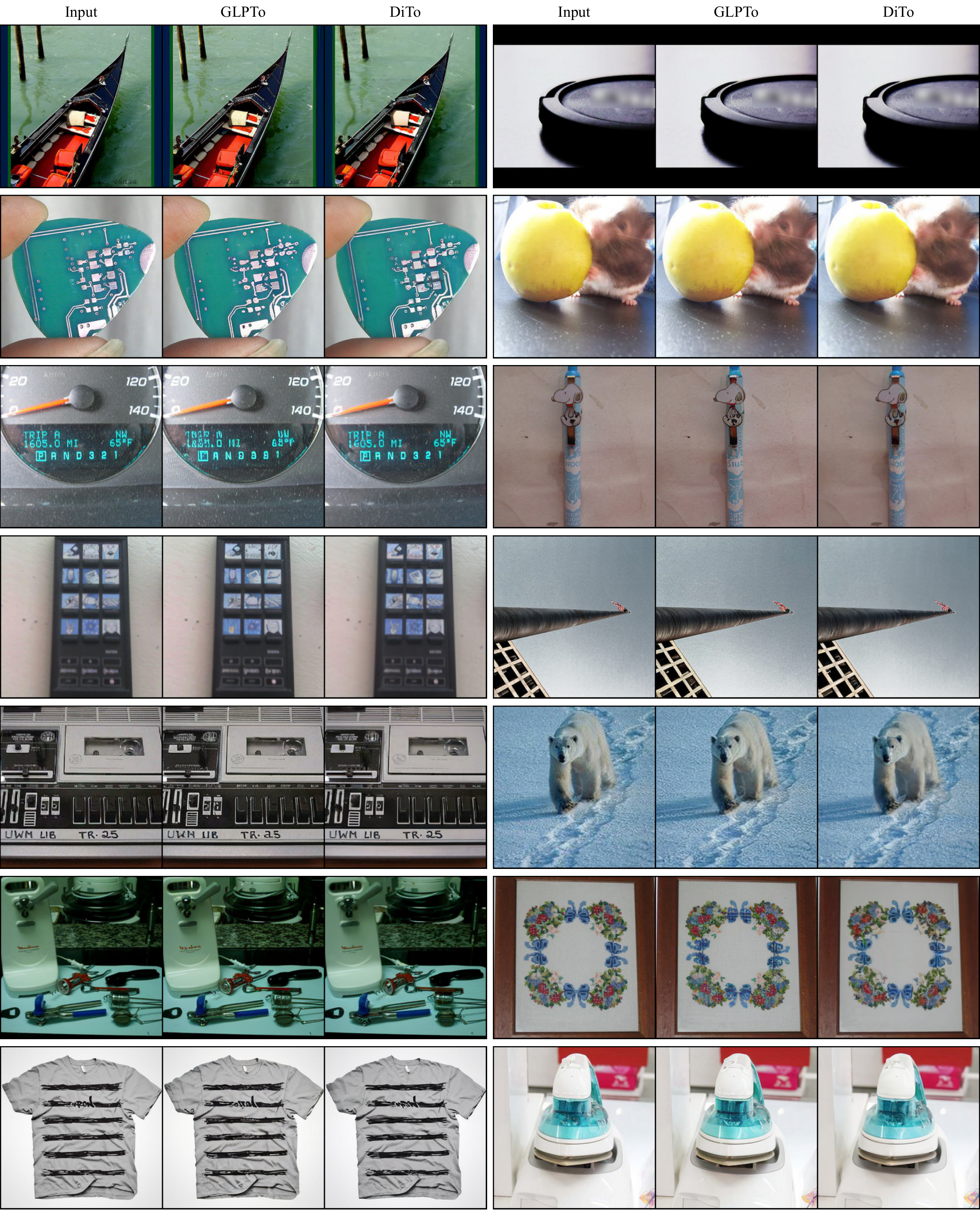}
    \caption{Additional qualitative comparison of tokenizers (at 256 pixel resolution).}
    \label{fig:supp_cmp_2}
\end{figure*}

We provide additional qualitative comparisons between the GAN-LPIPS tokenizer (\gae) and the diffusion tokenizer (\Ours) at XL size. The input images and corresponding reconstruction results are shown in \cref{fig:supp_cmp_1} and \cref{fig:supp_cmp_2}. We observe that \gae and \Ours are competitive in general, and \Ours usually has a better reconstruction quality for regular visual structures, symbols, and text.

\end{document}